%% file: ycviu-template-with-authorship.tex
\documentclass[times,twocolumn,final,authoryear]{elsarticle}

\usepackage{ycviu}
\usepackage{framed,multirow}

\usepackage{amssymb}
\usepackage{latexsym}

\usepackage{url}
\usepackage{xcolor}

\usepackage{multirow}
\usepackage{algorithm}
\usepackage{algorithmicx}
\usepackage{algpseudocode}
\usepackage{amsmath}
 \usepackage{array}
 \usepackage{comment}

\definecolor{newcolor}{rgb}{.8,.349,.1}

\newcommand{\highlight}[1]{{\textcolor{black}{#1}}}

\journal{journal}

\begin{document}

\thispagestyle{empty}

\clearpage

\ifpreprint
  \setcounter{page}{1}
\else
  \setcounter{page}{1}
\fi

\begin{frontmatter}

\title{Spoof Face Detection Via Semi-Supervised Adversarial Training}

\author[1]{Chengwei \snm{Chen}}
\author[1]{Wang \snm{Yuan}}
\author[2]{Xuequan \snm{Lu}\corref{cor1}} 
\cortext[cor1]{Corresponding author 
  %Tel.: +0-000-000-0000;  
  %fax: +0-000-000-0000;
  }
\ead{xuequan.lu@deakin.edu.au}
\author[1]{Lizhuang \snm{Ma}}

\address[1]{East China Normal University, North Zhongshan Road Campus: 3663 N, Shanghai and 200062, China}
\address[2]{Deakin University, 75 Pigdons Rd, Waurn Ponds VIC 3216, Australia}
%\address[3]{East China Normal University, North Zhongshan Road Campus: 3663 N, Shanghai and 200062, China}

%\address[4]{East China Normal University, North Zhongshan Road Campus: 3663 N, Shanghai and 200062, China}

\received{}
\finalform{}
\accepted{  }
\availableonline{  }
\communicated{}

\begin{abstract}
Face spoofing causes severe security threats in face recognition systems. Previous anti-spoofing works focused on supervised techniques, typically with either binary or auxiliary supervision. Most of them suffer from limited robustness and generalization, especially in the cross-dataset setting. In this paper,
\textcolor{black}{we propose a semi-supervised adversarial learning framework for spoof face detection, which largely relaxes the supervision condition. To capture the underlying structure of live faces data in latent representation space,} we propose to train the live face data only, with a convolutional Encoder-Decoder network acting as a Generator. Meanwhile, we add a second convolutional network serving as a Discriminator. The generator and discriminator are trained by competing with each other while collaborating to understand the underlying concept in the normal class(live faces). Since the spoof face detection is video based (i.e., temporal information), we intuitively take the optical flow maps converted from consecutive video frames as input. Our approach is free of the spoof faces, thus being robust and general to different types of spoof, even unknown spoof. Extensive experiments on intra- and cross-dataset tests show that our semi-supervised method achieves better or comparable results to state-of-the-art supervised techniques.
\end{abstract}

%\begin{keyword}
%\KWD Keyword1\sep Keyword2\sep Keyword3

%\end{keyword}

\end{frontmatter}

%\linenumbers

\input{paper/introduction.tex}

\input{paper/relatedwork.tex}

\input{paper/methodology.tex}

\input{paper/results.tex}

\input{paper/conclusion.tex}
%-----------------------------------------------------------------------

\bibliographystyle{model2-names}
\bibliography{refs}

\end{document}

%% file: paper/introduction.tex
\section{Introduction}
Biometrics plays a key part in authentication and security applications. Access control using face, fingerprint or iris has been existed for quite a while in our daily life. Face recognition, one of the prevalent biometric applications, has achieved noticeable successes \citep{galbally2014biometric}. \highlight{Face data has been a promising data type,  due to its convenience, universality and acceptability for users.  However, traditional face recognition systems can be easily fooled with common attacks like printed facial photographs. To obtain access to systems, criminals are already using some techniques to accurately simulate the biometric characteristics of valid users, such as faces.} This process is known as face spoofing attack, which poses a great threat to face recognition systems \citep{patel2016secure,ratha2001analysis}. Presentation attacks (abbreviated as PA), including printed paper face, replaying a video and wearing a mask, are one of the most prevalent face spoofs. It has been demonstrated that traditional face recognition systems could be vulnerable to PA \citep{chetty2006multi,frischholz2000biold,frischholz2003avoiding}. \highlight{Therefore, it is necessary to design robust countermeasure techniques to deal with the weakness in traditional face  recognition application and prevent such frauds. }As a result, various face anti-spoofing techniques have been proposed to detect spoof and live faces, before the face recognition stage. The main challenge of face anti-spoofing is how to achieve robustness and generalization to different kinds of PA. 

\highlight{Many previous strategies have been proposed to deal with the spoof face detection task. Spatial image information plays a critical role in face recognition system. Each facial region in our face includes different visual patterns and rich and discriminative information. These information could help to distinguish some faces from others. Therefore, some strategies are proposed  to find different spoofing cues from different facial regions by using handcrafted features, such as  LBP \citep{de2012lbp} and HOG \citep{yang2013face}. It is hard to obtain robust texture features, due to the cost of handcrafted features and a lack of an explicit correlation between pixel intensities and different types of attacks. With the recent development of deep learning, face spoofing detection based on high-level learning features achieve more promising performance. However, these CNN-based methods \citep{li2016original,patel2016cross} are adopted in spoof face detection with a softmax loss based on binary supervision. These supervised methods have the risk of overfitting on the training data and obtain low performance in the cross-dataset setting. In addition, temporal information is also a critical part in spoof face detection. For example, a liveness detection method \citep{bao2009liveness} is proposed for spoofing face detection with using optical flow. It attempts to find the differences in motion patterns. That model attempts to learn the concept of optical flow generated by 3D objects and 2D planes. The motion of an optical flow field consists of four basic movements: translation, rotation, moving, and swing. Previous motion based methods \citep{jee2006liveness,sun2007blinking,kollreider2005evaluating} usually need to learn or obtain some explicit features using complicated modules such as modeling the motion. Based on these features which focus on representing specific characteristics, the trained model can make the real face images and spoof face images more separable. However, because of the specificity, these methods are hard to be generalized to other spoofing types.}

Previous face anti-spoofing works focused on supervised methods, with the utilization of hand-crafted or learned features. Most approaches typically depend on binary or auxiliary supervision. Nevertheless, many previous works suffer from the following major limitations partially or wholly: (1) fully supervised setting--the utilization of both live and spoof face data (with labels), (2) the assumption of binary classification, and (3) the impracticality to take all types of spoof (maybe unknown spoof) into account. Furthermore, collecting spoof face data for training purpose is costly and time-consuming. Also, binary supervision could be insufficient to learn a good model and make desired predictions in cross dataset scenario. As a result, those face anti-spoofing techniques have limited robustness and generalization to various types of spoofing.

Motivated by the above limitations and analysis, we propose a novel adversarial network for anti-spoofing under the semi-supervised setting. We propose to train the live face data only, with a convolutional Encoder-Decoder network acting as a Generator. Besides, a second convolutional network is regarded as a Discriminator. The generator attempts to reconstruct the original input sample to fool the discriminator, while the discriminator tries to distinguish original images from generated images. In the process of training, both sub-networks compete with each other to achieve high-quality reconstructions for live faces data only.

While testing, the learned model has a lower reconstruction error of live face data than spoof face data. This is mainly because we train on live face data only, the model captures the real characteristic of live faces samples and the learned model can better describe the characteristics of live faces than those of spoof faces. We naturally take the optical flow maps converted from consecutive video frames as input, as the task of spoof face detection is video-based and involves temporal information. The semi-supervised setting significantly reduces the efforts in collecting spoof face data, thus making our method more robust and general to different types of face spoofing. As such, the proposed approach is practical in the real world. We validate our method on challenging datasets. We also compare our semi-supervised method with state-of-the-art supervised anti-spoofing techniques, showing that our method produces better or comparable results to those approaches.

In summary, the main contributions of this paper are:
\begin{itemize}
	\item a novel semi-supervised approach training on live face data only for spoof face detection.
	\item we propose a framework trained by generator and discriminator adversarially while collaborating to understand the real underlying concept in the normal class and classifying the testing samples by pixel-wise reconstruction error.
	\item we design a domain adaption algorithm which tries to learn some transfer components across domains in a Reproducing Kernel Hilbert Space (RKHS) using Maximum Mean Discrepancy (MMD).
	\item validation on challenging datasets, and extensive comparisons (intra- and cross-dataset testing) with current supervised anti-spoofing techniques.
\end{itemize}

\highlight{
The rest the paper is organized as follows. Section \ref{sec:relatedwork} reviews the relevant research. We elaborate our approach in Section \ref{sec:method}. Section \ref{sec:results} gives various experimental results, and Section \ref{sec:conclusion} concludes our work.
%There are four sections in this paper. The first section introduces the definition and statement of problem. In related work section, four types of previous methods are presented, such as traditional feature based methods, methods based on temporal information, hybrid methods and methods based on other cues. In methodology part, the details of proposed strategy and optimization method is illuminated. Moreover, both of the proposed loss functions are shown to optimize the framework in the training process. Trained model is adopted to distinguish the live faces from spoofing faces during the inference. To show the effectiveness of proposed method, the experiments part are divided into two subsections, ablation study and state-of-art comparison. In ablation study, we conduct the experiment with or without the discriminator part. In addition, to explore the influence of optical flow information, we consider to use appearance information and motion information in the experiments, respectively. Finally, intra-and cross-dataset experiments show that our semi-supervised method achieves better or comparable results to state-of-the-art supervised techniques.
}

%% file: paper/relatedwork.tex
\begin{table*}
\footnotesize
	\caption{\highlight{Characteristics of different face spoof detection methods.} }
	\label{differentmethods}
	\centering
		\begin{tabular}{|p{2cm}|p{3cm}|p{5cm}|p{4cm}|p{2cm}|}
			\hline
			\highlight{\textbf{Methods}} &\highlight{\textbf{Analysis type}} &\highlight{\textbf{Strategy}} &\highlight{\textbf{Datasets}} &\highlight{\textbf{Algorithm type}}\\
			\hline
			\highlight{LBP}  &	\highlight{Texture analysis} &	\highlight{Micro-texture analysis via LBP with SVM as a classifier} &	\highlight{NUAA Photograph Imposter Database} &
	\highlight{Supervised}  \\
		\hline
		
\highlight{DoG-SL} &	\highlight{Texture analysis} &
	\highlight{Applying an adaptive histogram equalisation to the images} &	\highlight{Yale Face Database and NUAA Photograph Imposter Database} &	\highlight{Supervised} \\
		\hline
		
\highlight{Color-texture} &	\highlight{Texture analysis} & \highlight{Computing a half of Face with another half that is divided in two ways: horizontally and vertically}&	\highlight{CASIA Face Anti-Spofing and the Replay-Attack databases} &
	\highlight{Supervised} \\
		\hline
		
\highlight{Optical flow field} &	\highlight{Motion analysis} & \highlight{Analyzing the optical flow field to detect real face}
&  \highlight{-}	&	\highlight{Supervised} \\
		\hline
	
\highlight{Structure tensor} &	\highlight{Motion analysis}	& \highlight{Face motion estimation based on the structure tensor and a few frames} &	\highlight{XM2VTS database} & 
	\highlight{Supervised} \\
		\hline
	
\highlight{Spatial-temporal Domain}  &	\highlight{Motion analysis + Texture analysis} &
	\highlight{A two-stream structure (spatial, temporal )} &
	\highlight{Replay-Attack, CASIA and 3DMAD} &
	\highlight{Supervised} \\
		\hline
	
\highlight{Patch-based CNN} &	\highlight{Texture analysis
+cue analysis} &
	 \highlight{Extracting the local features and holistic depth maps from the face images }
	& \highlight{CASIA-FASD, MSU-USSA, and Replay-Attack }&
	\highlight{Supervised} \\
		\hline
		
\highlight{VGG} &	\highlight{Texture analysis} &
	\highlight{Extracting the deep partial features from the convolutional neural network (CNN)} 
	& \highlight{Replay-Attack and CASIA} &
	\highlight{Supervised} \\
		\hline

\highlight{Liveness optical flow} &	\highlight{Motion analysis} &
	\highlight{Applying the Support Vector Machine to distinguish between the motion information of real
faces and photographs} 
	& \highlight{Regensburg university dataset} &
	\highlight{Supervised} \\

		\hline
\highlight{Auxiliary}  &	\highlight{Texture analysis
+ cues analysis} &
	\highlight{Fusing the estimated depth and rPPG  to distinguish live v.s. spoof faces}	& \highlight{CASIA-MFSD and Replay-Attack}  & 
	\highlight{Supervised} \\
	
		\hline
	\highlight{De-Spoof} &	\highlight{Texture analysis} &	\highlight{A CNN architecture with proper constraints and supervisions} &
	\highlight{Oulu-NPU, CASIA-MFSD and Replay-Attack} &	\highlight{Supervised} \\
	
		\hline
\highlight{Our method}	 &  \highlight{Motion analysis}	 & \highlight{A semi-supervised adversarial learning framework} &	\highlight{Nuaa, CASIA-MFSD and Replay-Attack} & 	\highlight{Semi-Supervised} \\
	
			\hline
		\end{tabular}
\end{table*}

\section{Related Work}
\label{sec:relatedwork}
The previous face anti-spoofing methods \citep{boulkenafet2017face,de2012lbp,de2013can,komulainen2013context,maatta2011face,mirjalili2017soft,patel2016secure,yang2013face} can be generally divided into four categories: feature based methods, temporal information based methods, Hybrid methods as well as approaches based on other cues. \highlight{Tab. \ref{differentmethods} compares the characteristics of these previous spoof detection methods, including LBP \citep{maatta2011face},
DoG-SL \citep{peixoto2011face},
Color-texture \citep{boulkenafet2015face},
Optical flow field \citep{bao2009liveness},
Liveness optical flow \citep{smiatacz2012liveness},
Structure-tensor\citep{kollreider2005evaluating},
Spatial-temporal domain \citep{sun2018investigation},
Patch-based CNN \citep{atoum2017face},
VGG \citep{li2016original},
Auxiliary \citep{liu2018learning} and
De-Spoof \citep{jourabloo2018face}. }

\subsection{Feature-based Methods}
Most early face anti-spoofing works used handcrafted features of texture information for binary classification (e.g., SVM). They expected that differing feature descriptors such as LBP \citep{de2012lbp,de2013can,maatta2011face}, HOG \citep{komulainen2013context,yang2013face}, DoG-SL \citep{komulainen2013context,yang2013face}, SIFT \citep{patel2016secure} and SURF \citep{chingovska2012effectiveness}  could be computed for live and spoof faces.
Nonetheless, many feature descriptors are largely affected by illumination, imagery and other factors. Such feature-based methods often have poor generalization in cross-dataset testing \citep{liu2018learning}.

CNN is good at extracting and learning deep features.  \citep{yang2014learn} treated CNN as a classifier for face anti-spoofing, and used different spatial scales of live and spoof face images for training.  \cite{xu2015learning} proposed a LSTM-CNN architecture to predict the frames of videos. Most previous CNN techniques for face anti-spoofing utilized a binary classification to predict live or spoof faces \citep{feng2016integration,li2016original,patel2016cross,yang2014learn}. However, both live and spoof face data have to be considered in the training procedure. In worse cases, the test face data does not involve cues like printed page edges or digital replay devices while the trained model might use such cues to detect spoof faces. As a result, the classification ability for live and spoof faces is limited. Also, it is difficult to explain the final results.

\begin{figure*}[h]
	\centering
	\includegraphics[width=\linewidth]{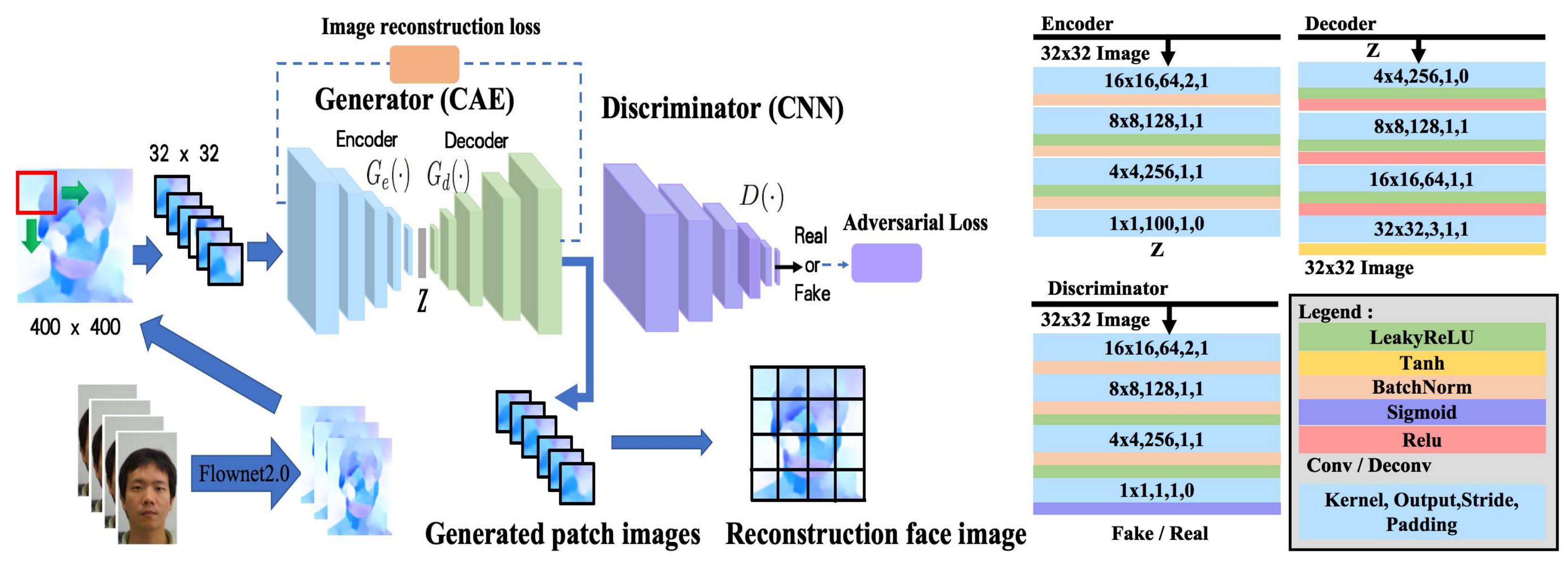}
	\caption{\textcolor{black}{Our framework consists of a generator and a discriminator. The generator and discriminator are trained by competing and collaborating with each other to understand the underlying structure in the live faces data. The architecture layers of each component are described on the right.
		}}
		%\Description{}
		\label{achieture}
	\end{figure*}

\subsection{Methods Based on Temporal Information}
As with other video-based tasks like activity recognition, temporal information is also useful in face anti-spoofing. Some researchers paid attention to the movement of key parts in a face, for example, eye-blinking and lip movements. The temporal information based methods are usually vulnerable to the replay attack (i.e., replaying video with a digital device).  \cite{gan20173d} proposed a 3D convolutional network to classify live and spoof faces, by supervisedly learning temporal features with a stacked structure. Unfortunately, it relies on a large amount of data and could perform poorly on small datasets.  \cite{xu2015learning} introduced a new structure by combining LSTM units with CNN for binarary classification.  \cite{feng2016integration} presented a CNN by taking both optical flow features and shearlet features as input. These methods took advantage of temporal information to distinguish between live and spoof faces.

\subsection{Hybrid Methods}
Hybrid techniques combining features and temporal information have also been proposed for spoof face detection.  \cite{schwartz2011face} used multiple low-level features to create one high dimensional vector with the size of more than one million. They further adopted the partial least squares approach on this vector to distinct between live and spoof faces.  \cite{komulainen2013complementary} introduced the combination of computationally inexpensive linear classifiers for robust face anti-spoofing. They used the fusion of motion information and features. Both methods depend on the multi-block local binary pattern and motion estimation from input videos.

\subsection{Methods Based on Other Cues}
There have been considerable amount of works using other cues derived from the original video frames \citep{komulainen2013context,george2019biometric}. For example, rPPG signal, IR image \citep{zhang2011face}, depth image \citep{wang2013face} and voice \citep{chetty2010biometric} are some common cues. Nevertheless, such cues have their own limitations. Taking rPPG-based methods as an instance, researchers often need to extract the rPPG signals from a long video, to achieve decent predictions \citep{liu2018learning}. As a matter of fact, it is unfeasible for a face anti-spoofing system to detect spoof faces through analyzing a long video (e.g., 50 seconds).

		\begin{figure*}[h]
			\centering
			\includegraphics[width=\linewidth,height=4cm]{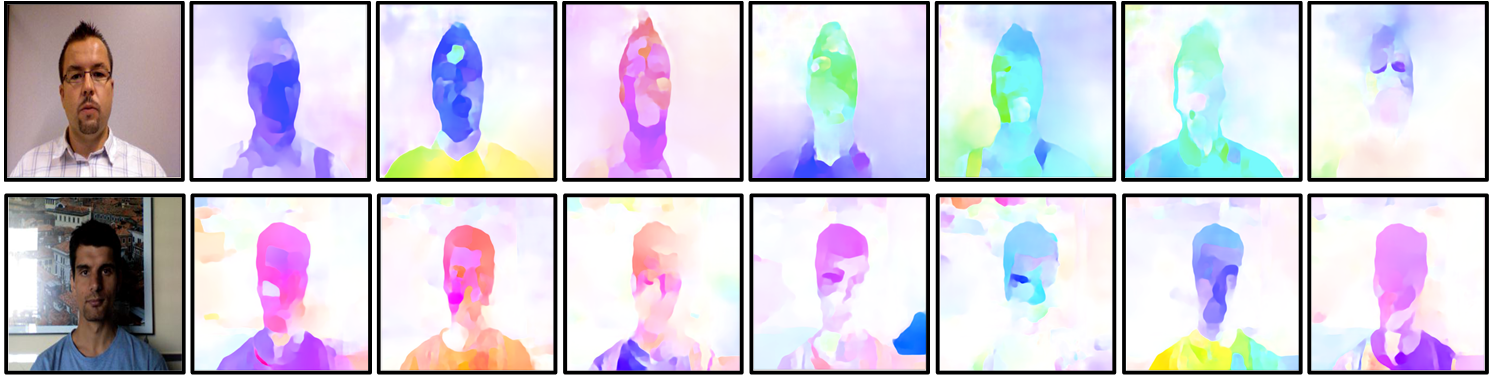}
			\caption{Two live faces with optical flow data visualization in each row. The first image is one of the frames in each live face video, followed by seven optical flow maps which are generated from its follow-up frames. }
			%\Description{}
			\label{live_optical_flow}
		\end{figure*}
		
		\begin{figure*}[h]
			\centering
			\includegraphics[width=\linewidth,height=4cm]{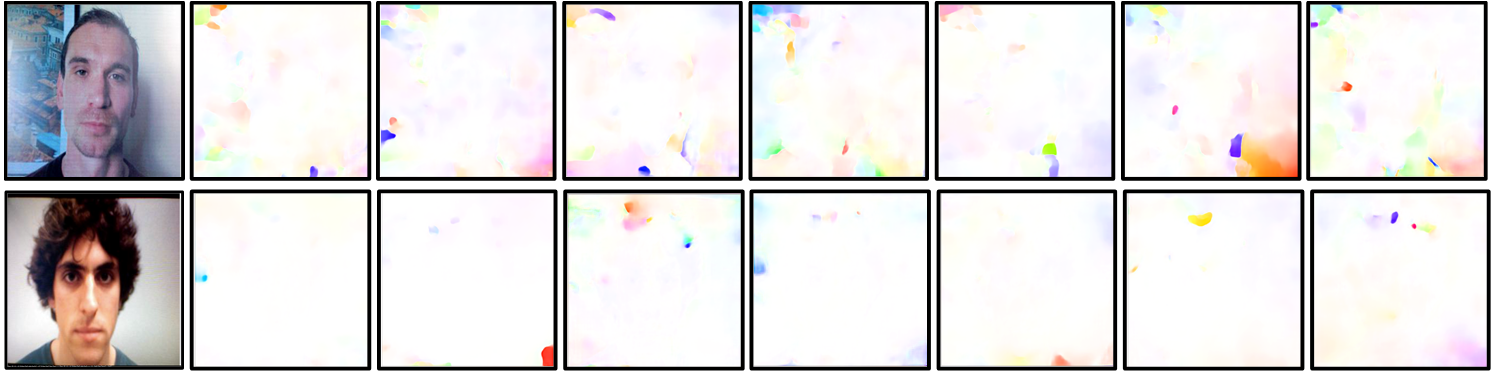}
			\caption{Two fixed spoofing faces with optical flow data visualization in each row. The first image is one of the frames in each spoofing face video by holding the client biometry, followed by seven optical flow maps which are generated from the follow-up frames. }
			%  \Description{}
			\label{fixed_optical_flow}
		\end{figure*}
		
		\begin{figure*}[h]
			\centering
			\includegraphics[width=\linewidth,height=4cm]{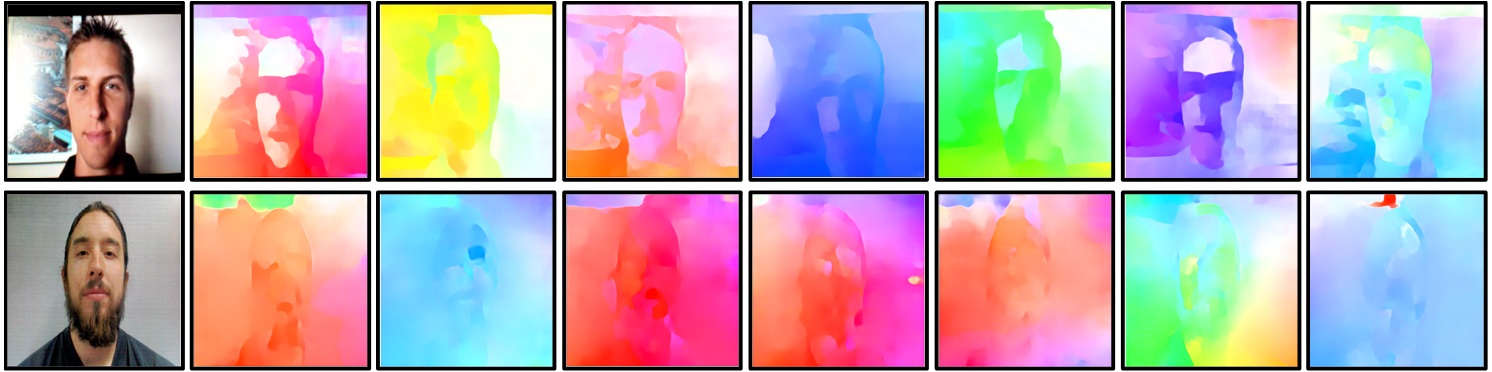}
			\caption{Two hand spoofing faces with optical flow data visualization in each row. The first image is one of the frames in each spoofing face video from the device held by the attacker's hands, followed by seven optical flow maps which are generated from the follow-up frames. }
			%  \Description{}
			\label{hand_optical_flow}
		\end{figure*}

\subsection{Anomaly detection}
	\highlight{Anomaly detection is a classical problem in computer vision. When samples are deviating from the expected behavior defined by ``normal'' samples of a training dataset, these samples are classified as the abnormal class. 
	%In the past, due to the power of sparse representation and dictionary learning, researchers use sparse representation to learn the dictionary of normal behaviors. In the process of testing, the patterns which have large reconstruction errors are considered as anomalous behaviors. 
	}
	
	% deep learning method for Anomaly detection and drawback
\highlight{Recently, deep learning based autoencoders are used to learn the pattern of normal behaviors and exploit the reconstruction loss to detect anomalies. For example, \cite{baur2018deep} tackles the problem by learning a mapping to a lower dimensional representation, where the real distribution is modeled. The decoder upscales the latent feature vector to reconstruct the image. In recent research, a lot of abnormal detection methods \citep{zenati2018adversarially,
xia2019latent} based on the Generative Adversarial Networks (GANs) are proposed. %The GAN style architecture is adversarially trained under the semi-supervised learning framework, in which the underlying structure of training data is captured in the latent space. In the process of testing, the abnormal samples are regarded as the out-of-distributions samples that naturally exhibit a higher pixel-wise reconstruction error than normal samples. 
For instance, \cite{xia2019latent} proposed latent spatial features based on generative adversarial networks for face anti-spoofing with an additional feature classifier. The input of this framework extracts the appearance information from the original face image with different sizes. In our work, instead we use the motion information from the original face images. According to the ablation study (Section \ref{sec:4.2.2}), the performance of using motion information is better than using appearance information. Moreover, each size of the input corresponds to one GAN model, which induces significantly higher costs. In addition, their framework only reports a high performance in the intra-dataset setting. It disregards the generalization issue by excluding the cross-dataset setting, which is critical for spoofing detection. }

%% file: paper/methodology.tex
\section{Proposed Approach}
\label{sec:method}
	
In this section, we present how to learn the intrinsic structure of live faces by using the proposed adversarial training framework. We start by describing the details of the overview network architecture, then depict each term in loss function, and finally give the description of the testing method.
	
	\subsection{Network Architecture}
Our method consists of a data preprocessing step and a GAN-style architecture. The preprocessing step is to convert consecutive video frames into optical flow maps. Fig. \ref{live_optical_flow},\ref{fixed_optical_flow},\ref{hand_optical_flow} shows the visualization of optical flow map. The GAN-style architecture, inspired by the anomaly detection \citep{sabokrou2018adversarially}, comprises of two components: the generation network and the discrimination network. \highlight{Fig. \ref{achieture} shows the overview of our framework.}

Due to the outstanding performance of CNN \citep{ krizhevsky2012imagenet,lawrence1997face,kalchbrenner2014convolutional}, we take a convolutional autoencoder as the Generator. The main idea is that we only consider the live face data for training. The learned model is therefore not good at depicting the characteristics of spoof face data, leading to high reconstruction errors. The reason why we employ Convolutional AutoEncoder (CAE) in the proposed framework can be concluded as follows: (1) Conventional Autoencoders (AEs) often ignore the structure of 2D images, and interpret the input as a single latent vector. (2) The network is constrained by the number of input images. The redundant parameters in AEs force each feature to be global by spanning the entire visual field. (3) The Convolutional AutoEncoder (CAE) can learn the optimal filters to minimize the reconstruction error. In fact, Convolutional Neural Networks are usually referred to supervised learning algorithms. CAE, instead, is trained only to learn filters to extract features that can be used to reconstruct the input. 
	
		\begin{figure}[b]
			\centering
			\includegraphics[width=\linewidth]{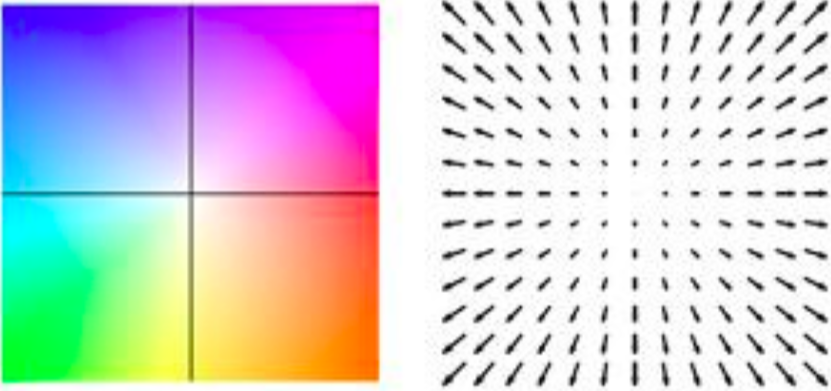}
			\caption{\highlight{Visualization of horizontal and vertical displacements in one RGB image. }}
			\label{Color}
		\end{figure}

	To prevent being fooled by the generator, the discriminator learns the core characteristics in the original data during the period of training. The discriminator also assists the generator to get robust and stable parameters in the process of training. This part of parameters would increase the reconstruction gap between live faces and fake faces in the process of testing.

	\subsection{Overall Loss Function}
	\textcolor{black}{To train our model, we define a loss function in Eq. \eqref{loss-function}  including two components, the adversarial loss and the pixel-wise image reconstruction loss.} 
	
	\begin{equation}\label{loss-function}
	\mathcal{L}=w_{i} \mathcal{L}_{irec} + w_{a} \mathcal{L}_{adv},
	\end{equation}
	where $w_{i}$ and $w_{a}$ are the weighting parameters balancing the impact of individual item to the overall object function.

\textbf{Adversarial loss.} The Generative Adversarial Network (GAN) \citep{ creswell2018generative} originates from a game between two players. One player is called the generator $G(x)$. The generator creates samples that are intended to come from the same distribution as the training data. The other player is called the discriminator $D(x)$. The discriminator would make a decision whether the samples are generated by the generator or taken from the training data. The generator attends to fool the discriminator by reconstructing fake samples similar to the true training data. This adversarial game between the generator and discriminator can be formulated as: 
	
	\textcolor{black}{
		\begin{equation}
		\label{eq:1}
		\begin{aligned} \mathcal{L}_{a d v}=& \min _{G} \max _{D}\left(E_{\boldsymbol{x} \sim p_{\mathbf{x}}}[\log (D(\mathbf{x}))]\right.\\ &\left.+E_{\boldsymbol{x} \sim p_{\mathbf{x}}}[\log (1-D(G(\mathbf{x})))]\right). \end{aligned}
		\end{equation} }

\textbf{Image reconstruction loss: } While the discriminator tries to differentiate between realistic images and generated images, and the generator trying to fool the discriminator. However, the generator is not optimized towards learning the real concept from input data only by adversarial loss. Some prior works have proposed that the distance between input images and generated images should be considered. \cite{isola2017image} shows that  the use of L1 yields less blurry results than L2. Therefore, we use L1 loss function to penalize the generator by minimizing the distance between original input $\mathbf{x}$ and generated images $G(x)$ as follows.

	\begin{equation}
	\label{eq:3}
	\mathcal{L}_{irec}=\mathbb{E}_{x \sim p_{\mathbf{x}}}\|x-G(x)\|_{1}
	\end{equation}

\subsection{Data Preprocessing}
\highlight{We extract frames from each video with 30 frames per second. FlowNet2.0 \citep{Ilg2017FlowNet2E} is then employed to estimate the optical flow between frames, due to its effectiveness. Optical flow is the pattern of apparent motion, which is calculated based on two adjacent images. It defines both horizontal and vertical displacements for each pixel, and reflects motion about objects and scene. The pre-trained FlowNet model estimates the optical flow between each pair of two adjacent frames and outputs the optical flow files. The horizontal and vertical components are included in optical flow files. The color-coding scheme \citep{lopez2017accurate} allows us to visualize the horizontal and vertical displacements in one image, as illustrated in Fig. \ref{Color}. Colors can be assigned to each pixel. %without interference due to noticeable displacements (motion). 
We utilize the color coding scheme to convert these optical flow files into images where the displacement vector is color. }

\highlight{The output flow maps are also RGB images with colors indicating the flow signal. The patches are generated from each flow map by a sliding window. The size of this window is set to 32 $\times$ 32.} Fig. \ref{live_optical_flow}, \ref{fixed_optical_flow} and \ref{hand_optical_flow} visualize optical flows of live faces, spoofing faces by holding the client biometry (i.e., fixed spoofing) and spoofing faces from the device held by the attacker's hands (i.e., hand spoofing), respectively. It shows that the optical flows of live faces are more clear than the spoof faces. Hand spoofing leads to considerable amount of noise on the flow maps. This is because that the movement of spoofing faces and digital device screens is consistent. For fixed spoofing faces, printed faces are fixed in front of the detection systems. This type involves few noise and nearly no optical flows.

\subsection{Testing method}
\label{sec:testing}
To demonstrate the effectiveness of the proposed framework, we conduct two intra-testing experiments and one cross-testing experiment. For intra-testing experiments, the model is trained in the training dataset accordingly, as with the state-of-the-art methods \citep{yu2017anisotropic}. The testing dataset in the same domain is used to evaluate the performance of each method. Different from intra-testing experiments, cross-database experiments with different domains are more challenging. Domain adaptation \citep{ finkel2009hierarchical} is a field associated with machine learning and transfer learning. The aim of the domain adaptation problem is to train a well performing model from the source data distribution. The trained model could still perform well on a different (but related) target data distribution. As such, we attempt to extend the domain adaptation in our study. \textit{To our knowledge}, we are the first to investigate the domain adaptation issue in the face anti-spoofing area.
	
In the cross-database situation, the labels of all target samples are unknown during training. Compared with the intra-database setting, it is more ubiquitous in real-world applications. Due to the unavailability of labels in the target domain, one commonly used strategy is to learn domain-invariant representations via minimizing the domain distribution discrepancy. In our cross-database scenario, the model is trained on dataset A and tested on dataset B. There exists some difference between the source domain and target domain, for example, image quality, reflection and environment. One intuitive solution is to consider mapping the reconstruction data to a high (possibly infinite) dimensional space and computing the sample means in this space using high-order statistics (up to infinity). As a result, we could achieve a better discrimination threshold for live and spoofing faces. By contrast, directly training a classifier on the source data and using the threshold set in the source data often leads to certain ``overfitting'' to the source distribution and reduced performance while testing on the target domain.
	
	We consider a source domain $\mathcal{D}_{s}=\{\boldsymbol{x}_{i}^{s}, y_{i}^{s}\}_{i=1, \ldots, n_{s}} $ and a target domain $\mathcal{D}_{t}=\{\boldsymbol{x}_{i}^{t}, y_{i}^{t}\}_{i=1, \ldots, n_{t}}$. Here, $\boldsymbol{x}_{i}^{s} \in \mathbb{R}^{N_{s}}, \boldsymbol{x}_{i}^{t} \in \mathbb{R}^{N_{t}}$ are the reconstruction errors for each frame in the source domain and the target domain, respectively. $y_{i}^{s} \in \mathcal{C}, y_{i}^{t} \in \mathcal{C}$ are corresponding labels, where the target labels $\left\{y_{i}^{t}\right\}_{i=1, \dots, n_{t}}$ are not available for training. For domain adaption, we assume that the source and target domains are associated with the same label space, while $\mathcal{D}_{s}$ and $\mathcal{D}_{t}$ are drawn from distributions $\mathbb{P}_{s} \text { and } \mathbb{P}_{t}$ which are assumed to be different. That is, the source and target distribution have different joint distributions of data $X$ and labels $Y$: $\mathbb{P}_{s}(\boldsymbol{X}, \boldsymbol{Y})$ $\neq \mathbb{P}_{t}(\boldsymbol{X}, \boldsymbol{Y})$.
	
	Maximum Mean Discrepancy (MMD) \cite{yan2017mind} is an effective non-parametric metric for comparing the distance between two distributions. Given two distributions $s$ and $t$, by mapping the data to a reproduced kernel Hilbert space (RKHS) using function $\phi(\cdot)$, the MMD between $s$ and $t$ is defined as,
	\begin{equation}
	\label{eq:3}
	\operatorname{MMD}(s, t)=\sup _{\|\phi\| \mathcal{_H} \leq 1}\|E_{\mathbf{x}^{s} \sim s}[\phi(\mathbf{x}^{s})]-E_{\mathbf{x}^{t} \sim t}[\phi(\mathbf{x}^{t})]\|_{\mathcal{H}},
	\end{equation}
	where $E_{\mathbf{x}^{s} \sim s}[\cdot]$ denotes the expectation with regard to the distribution $s$, and $\|\phi\|_{\mathcal{H}} \leq 1 $  defines a set of functions in the unit ball of a RKHS. Based on the statistical tests defined by MMD, we have $\text{MMD}(s, t) = 0 \iff s = t$. Denote by $\mathcal{D}_{s}=\left\{\mathbf{x}_{i}^{s}\right\}_{i=1}^{M}$ and $\mathcal{D}_{t}=\left\{\mathbf{x}_{i}^{t}\right\}_{i=1}^{N}$, two sets of samples drawn i.i.d. from the distributions $s$ and $t$ respectively, the empirical estimation of MMD can be given by: 
	\begin{equation}
	\label{eq:4}
	\operatorname{MMD}\left(\mathcal{D}_{s}, \mathcal{D}_{t}\right)=\left\|\frac{1}{M} \sum_{i=1}^{M} \phi\left(\mathbf{x}_{i}^{s}\right)-\frac{1}{N} \sum_{j=1}^{N} \phi\left(\mathbf{x}_{j}^{t}\right)\right\|_{\mathcal{H}},
	\end{equation}
	where $\phi(\cdot)$ denotes the feature map associated with the kernel map $
	k\left(\mathbf{x}^{s}, \mathbf{x}^{t}\right)=\left\langle\phi\left(\mathbf{x}^{s}\right), \phi\left(\mathbf{x}^{t}\right)\right\rangle$, which is usually defined as the convex combination of several basis kernels.

	With the help of MMD, the statistical test method works in the following way. Based on the samples of two distributions, one distribution is the reference distribution formed by training live face samples, and another distribution is obtained in the same way from test samples. By finding the continuous function $\phi$ in the sample space, the mean value of the samples from different distributions on function $\phi$ is obtained. Dividing the two mean values yields an average difference between the two distributions. Finally, MMD is taken as the measurement to determine the category of the test videos. If the value of MMD is smaller than the predefined threshold T, the test samples distribution is considered to be the close to the live face reference distribution; otherwise they are spoof videos. The final testing scheme is summarized in Algorithm 1.

	\floatname{algorithm}{Algorithm}
	\renewcommand{\algorithmicrequire}{\textbf{input:}}
	\renewcommand{\algorithmicensure}{\textbf{output:}}
	\begin{algorithm}
		\footnotesize
		\caption{Spoofing Detection in video V}
		\begin{algorithmic}[1] %每行显示行号
			\Require A video $V=[F_1,F_2,...F_N]$, trained models: Generator, Auxiliary encoder and decision threshold T
			\Ensure report if (ScoreVideo $>$ T) 'spoof' else 'non-spoof'
			\Function{Distribution}{$V$} 
			\State FrameArray=[]
			\For{$k \gets 1$ to $N$}
			\State ScoreFrame=$\left\|G_{e}(F_k)-G_{e^{\prime}}(F_k^{\prime})\right\|_{1}$
			\State FrameArray $\gets$ FrameArray +ScoreFrame
			\EndFor
			\State \Return FrameArray
			\EndFunction
			
			\State ReferDis= Distribution(TrainPosVideo)
			\State TestDis= Distribution(TestVideo)
			\State ScoreVideo = MMD(ReferDis,TestDis)
		\end{algorithmic}
	\end{algorithm}

%% file: paper/results.tex
\section{Experimental Results}
\label{sec:results}
In this section, we firstly present the experimental setting, including datasets, and more implementation details. Then, for ablation study, two experiments are conducted to analyze the proposed method in detail. Finally, We evaluate proposed method and the state-of-the-art techniques on both intra/inter-dataset settings.

\begin{table*}[ht]
	\scriptsize
	\centering 
	\renewcommand\tabcolsep{11.0pt} 
	\caption{\highlight{Abnormal detection results for MNIST/CIFAR10 datasets using Protocol 2. (Plane and Car classes are annotated as Airplane and Automobile in CIFAR10). }}
	\label{sec:toy}
	\begin{tabular}{c|c|c|c|c|c|c|c|c|c|c|c}
		\hline 
		&\textbf{0} &\textbf{1}&\textbf{2}&\textbf{3}&\textbf{4}&\textbf{5}&\textbf{6}&\textbf{7}&\textbf{8}&\textbf{9}&\textbf{MEAN}\\
		\hline 
		\textbf{OCSVM  ('01)}&0.988&0.999&0.902&0.950&0.955&0.968&0.978&0.965&0.853&0.955&0.9513\\
		
		\textbf{KDE  ('06)}&0.885&0.996&0.710&0.693&0.844&0.776&0.861&0.884&0.669&0.825&0.8143\\
		
		\textbf{DAE  ('06)}&0.894&0.999&0.792&0.851&0.888&0.819&0.944&0.922&0.740&0.917&0.8766\\
		
		\textbf{VAE  ('13)}&0.997&0.999&0.936&0.959&0.973&0.964&0.993&0.976&0.923&0.976&0.9696\\
		
		\textbf{Pix CNN  ('16)}&0.531&0.995&0.476&0.517&0.739&0.542&0.592&0.789&0.340&0.662&0.6183\\
		
		\textbf{AND  ('19)}&0.984&0.995&0.947&0.952&0.960&0.971&0.991&0.970&0.922&0.979&0.9671\\
		
		\textbf{DSVDD  ('18)}&0.980&0.997&0.917&0.919&0.949&0.885&0.983&0.946&0.939&0.965&0.9480\\
		
		\textbf{Autoencoder}
		&0.992&\textbf{1.0}	&0.876&0.937&0.949&0.968&0.984&0.959&0.843&	0.959	&0.9467\\
		
		\textbf{Proposed method}&\textbf{0.996}	&0.999&\textbf{0.987}&\textbf{0.986}&\textbf{0.977}&	\textbf{0.991}&	\textbf{0.998}&	\textbf{0.987}	&\textbf{0.986}&	\textbf{0.987}&	\textbf{0.9898}\\
		
		\hline 
		\hline 
		&\textbf{PLANE} &\textbf{CAR}&\textbf{BIRD}&\textbf{CAT}&\textbf{DEER}&\textbf{DOG}&\textbf{FROG}&\textbf{HORSE}&\textbf{SHIP}&\textbf{TRUCK}&\textbf{MEAN}\\
		\hline 
		\textbf{OCSVM ('01)}&0.630 &0.440 &0.649 &0.487& 0.735& 0.500 &0.725& 0.533& 0.649& 0.508& 0.5856\\
		
		\textbf{KDE  ('06)}&0.658&0.520&0.657&0.497&0.727&0.496&0.758&0.564&0.680&0.540&0.6097\\
		
		\textbf{DAE  ('06)}&0.411&0.478&0.616&0.562&0.728&0.513&0.688&0.497&0.487&0.378&0.5358\\
		
		\textbf{VAE ('13)}&0.700&0.386&0.679&0.535&0.748&0.523&0.687&0.493&0.696&0.386&0.5833\\
		
		\textbf{Pix CNN  ('16)}&0.788&0.428&0.617&0.574&0.511&0.571&0.422&0.454&0.715&0.426&0.5506\\
		
		\textbf{AND ('19)}&0.717&0.494&0.662&0.527&0.736&0.504&0.726&0.560&0.680&0.566&0.6172\\
		
		\textbf{DSVDD  ('18)}&0.617&\textbf{0.659}&0.508&0.591&0.609&0.657&0.677&\textbf{0.673}&0.759&\textbf{0.731}&0.6481\\
		
		\textbf{Autoencoder}&0.735	&0.585&\textbf{0.752}&0.703&0.375&0.687&0.594&	0.397	&0.781&	0.500&0.6109\\
		\textbf{Proposed method}&\textbf{0.996}	&0.648&\textbf{0.752}&\textbf{0.770}&\textbf{0.934}&	\textbf{0.695}&	\textbf{0.958}&	0.623	&\textbf{0.976}&	0.587&	\textbf{0.7930}\\
		\hline 	
	\end{tabular}	
\end{table*}

\subsection{Experimental Setup and Datasets}
The proposed method is mainly implemented in the Tensorflow framework \citep{tensorflow2015-whitepaper}. The experiments are carried out on a PC with a NVIDIA-1080  graphics card and a multi-core 2.1 GHz CPU. A good face anti-spoofing system must be robust to different types of attacks. We evaluate our method and the state-of-the-art techniques on three publicly available face spoofing detection databases: (i) NUAA Imposter Database \citep{ tan2010face}, (ii) Replay-Attack \citep{ chingovska2012effectiveness} dataset, and (iii) CASIA MFSD \citep{ zhang2012face} dataset. These structures are kept fixed for all databases, and learning rate is set to $0.02$.

\begin{figure}[h]
	\centering
	\includegraphics[width=\linewidth,height=3cm]{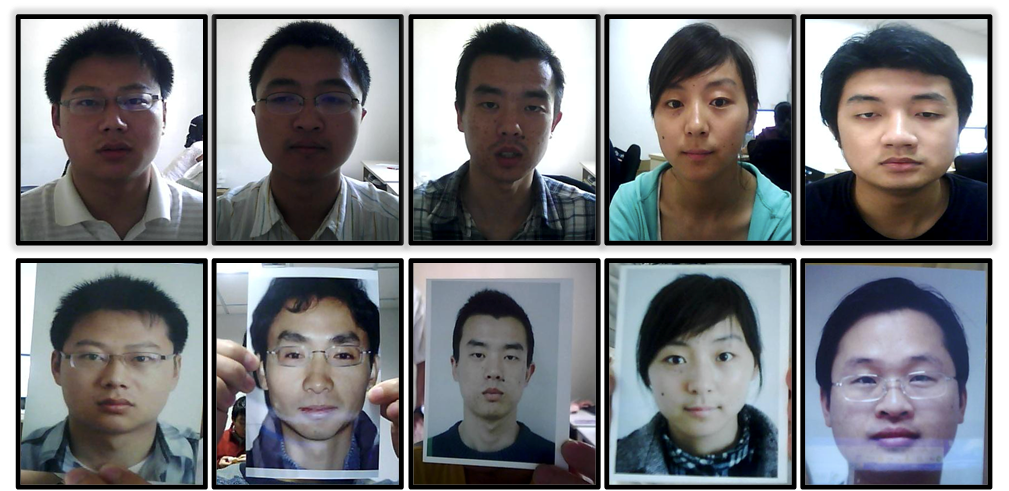}
	\caption{Some samples from the NUAA dataset. The first row and second row show five live samples and five spoofing samples, respectively. }
	%\Description{}
	\label{NUAA_sample}
\end{figure}

According to the work \citep{akcay2018ganomaly}, CIFAR10 and MNIST datasets are used to construct the experiment to illustrate the superiority of our approach over the state-of-the-art one-class classifiers. One of the classes is regarded as normal class, while the rest ones belong to the abnormal class. In particular, we respectively get ten sets for MNIST and CIFAR10, and then detect the outlier anomalies by only training the model on the normal class data in ablation study. 

The NUAA dataset is widely used for the evaluation of face liveness detection. This dataset consists of 15 different subjects captured in different places and illumination conditions, involving $12,614$ real and photographed face images. Each subject was asked to look at the webcam frontally with a neutral expression and without noticeable movements such as eyeblink or head movement. For training data, it contains $1,743$ real faces and $1,748$ photographed faces. For testing, it includes $3,362$ real faces and $5,761$ photographed faces. Fig. \ref{NUAA_sample} shows some samples from the NUAA dataset.

\begin{figure}[h]
	\centering
	\includegraphics[width=\linewidth,height=3cm]{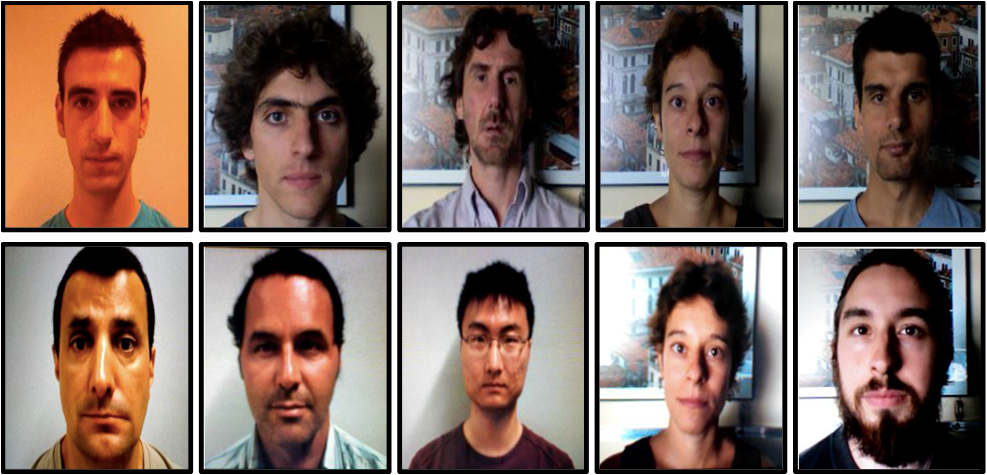}
	\caption{Some samples from the Replay-Attack dataset. It includes some live faces, some spoofing faces by holding the client biometry (i.e., fixed spoofing) and spoofing faces from the device held by the attacker's hands (i.e., hand spoofing).
	}
	% \Description{}
	\label{replay_samples}
\end{figure}

The Replay-Attack dataset is also a widely used and publicly available database. It has $360$ videos ($60$ real faces videos and $300$ spoof faces) as the training data.  About validation data, it has the same number of videos as training videos. The validation data will be fully used, to calibrate the threshold to distinguish between real and spoof faces (explained in Section 3.4). The resolution of the Replay-Attack data is $320 \times 240$. The dataset considers different lighting conditions used in spoofing attacks. It consists of $80$ videos of real faces and $400$ videos of fake faces as the testing data. The fake faces are obtained by using the attackers' bare hands or fixed support. Fig. \ref{replay_samples} shows some samples from the Replay-Attack dataset.

The CASIA \citep{ zhang2012face} dataset involves $50$ subjects, and each subject has $12$ videos ($3$ real faces and $9$ fake faces). The dataset is divided into the training set ($20$ subjects, $240$ videos) and the test set ($30$ subjects, $360$ videos). Compared with the Replay-Attack dataset, there is no validation data in this dataset. The CASIA dataset is more difficult in spoof face detection, in terms of image quality, resolution and video length. It consists of print and replay attacks using corresponding photos and replayed videos. Some of the print attack photos are manually cropped around the eyes to deter eye-blinking based techniques.

There exist a lot of face anti-spoofing approaches, and many CNN-based supervised works have achieved promising results in the intra-database setting. However, the high intra-database prediction accuracy does not guarantee a decent performance in the inter-database setting which is more common in real world. In fact, the cross-database performance better reflects the actual capability of a system in real-world applications. Therefore, a good cross-database performance provides strong evidence that: i) features are generally invariant to different scenarios (i.e., camera and illuminations), ii) a spoof classifier trained in one scenario is generalizable to other scenarios, and iii) data captured in one scenario can be useful for developing effective spoof detectors in other scenarios. As such, to demonstrate the effectiveness of the proposed framework, we conduct two intra-database experiments and two cross-database experiments.

\subsection{Ablation study}
%\highlight{In this section, we analysis proposed method in different perspectives. The first part is to explore the the effects with/ without discriminator part in proposed method. The second part is to show the impact on the performance under the different types of input.}

\subsubsection{\highlight{With/without the discriminator}}

\highlight{
To show the effectiveness of adversarial learning, it is necessary to conduct the experiment with or without the discriminator part.
In the first scenario, we only use the common convolutional autoencoder with a simple image reconstruction error. During the inference, the reconstruction error of the test sample is regarded as the abnormality score. In the second scenario, all components are used in the proposed framework with the discriminator. Besides the image reconstruction error, the adversarial learning loss is also considered. The generator tries to generate a high quality image to fool the discriminator. The discriminator attempts to distinguish the generated image from a realistic image. During the training process, the discriminator helps the generator to capture the underlying concept of normal samples. The test sample is detected in the same way as the first scenario (i.e., image reconstruction error). }

\highlight{Since we formulate the spoof face detection task as abnormal detection task, it is necessary to explore the effectiveness of the proposed method in both tasks. }
\begin{table}[h]
	\scriptsize
	\caption{\highlight{Classification performance of autoencoder (without discriminator) and the proposed method (with discriminator), in terms of HTER (\%). They are trained using the CASIA-MFSD dataset and tested on the Replay-Attack dataset, and vice versa.} }
	\label{ablationcrossdataset}
	\begin{tabular}{c|c|c|c|c|c}
		\hline
		\multicolumn{1}{c|}{\multirow{2}{*}{\textbf{Methods}}} & \multicolumn{1}{l|}{\textbf{Train}}                                                & \textbf{Test}& \multicolumn{1}{l|}{\textbf{Train}} & \textbf{Test}                                                    & \multicolumn{1}{|c}{\multirow{2}{*}{\textbf{Average}}} \\ \cline{2-5}
		& \multicolumn{1}{l|}{\begin{tabular}[c]{@{}l@{}}\textbf{CASIA}\\\textbf{MFSD}\end{tabular}} & \begin{tabular}[c]{@{}l@{}}\textbf{Replay}\\\textbf{Attack}\end{tabular} &
		\multicolumn{1}{l|}{\begin{tabular}[c]{@{}l@{}}\textbf{Replay}\\\textbf{Attack} \end{tabular}} & \begin{tabular}[c]{@{}l@{}} \textbf{CASIA}\\\textbf{MFSD}    \end{tabular} & \multicolumn{1}{|c}{}                         \\ \hline
		\textbf{	Autoencoder    }         & \multicolumn{2}{c|}{25.6\%} & \multicolumn{2}{c|}{47.3\%} &36.5\%                            \\ 
		\textbf{	Proposed method   }     & \multicolumn{2}{c|}{15.6\%} & \multicolumn{2}{c|}{44.1\%}&  29.8\%                           \\ \hline
	\end{tabular}
\end{table}
\highlight{In the abnormal detection task, we conduct an experiment to demonstrate the superiority of our method over state-of-the-art one-class classifiers on MNIST and CIFAR10 datasets, shown in Tab. \ref{sec:toy}. For both MNIST and CIFAR10, we select one class as the normal class at each time, while leaving the rest to be the abnormal classes, leading to ten sets for abnormal detection. Normal data and abnormal data are to imitate live faces and spoof faces, respectively. Our method typically achieves improvements compared with other methods, including OCSVM \citep{scholkopf2001estimating}, KDE  \citep{bishop2006pattern}, DAE \citep{hadsell2006dimensionality}, VAE \citep{kingma2013auto}, Pix CNN \citep{kalchbrenner2016conditional},  AND \citep{abati2019latent} and DSVDD  \citep{ruff2018deep}. In addition, it is clear that the proposed method with the discriminator achieves higher performance than the autoencoder without the support from discriminator in both datasets. In our spoofing face detection task, the cross-dataset experiment is also conducted by using the two scenarios (with and without the discriminator) described above. As shown in Tab. \ref{ablationcrossdataset}, the discriminator helps the generator (Autoencoder) to capture the concept of live faces. The trained generator is used to detect the spoof faces directly, and we obtain a better performance with the dsicriminator than the autoencoder strategy without the discriminator. }

\subsubsection{\highlight{Impact on performance with optical flow }}
\label{sec:4.2.2}
									
\begin{table}[h]
	\scriptsize
	\caption{\highlight{Classification performance of the proposed approach in different types (motion or appearance) of input,  in terms of HTER (\%). The algorithm is trained using the CASIA-MFSD dataset and tested on the Replay-Attack dataset, and vice versa.}}
	\label{crossdataset_ab}
	\begin{tabular}{c|c|c|c|c|c}
		\hline
		\multicolumn{1}{c|}{\multirow{2}{*}{\textbf{Methods}}} & \multicolumn{1}{l|}{\textbf{Train}}                                                & \textbf{Test}& \multicolumn{1}{l|}{\textbf{Train}} & \textbf{Test}                                                    & \multicolumn{1}{|c}{\multirow{2}{*}{\textbf{Average}}} \\ \cline{2-5}
		& \multicolumn{1}{l|}{\begin{tabular}[c]{@{}l@{}}\textbf{CASIA}\\\textbf{MFSD}\end{tabular}} & \begin{tabular}[c]{@{}l@{}}\textbf{Replay}\\\textbf{Attack}\end{tabular} &
		\multicolumn{1}{l|}{\begin{tabular}[c]{@{}l@{}}\textbf{Replay}\\\textbf{Attack} \end{tabular}} & \begin{tabular}[c]{@{}l@{}} \textbf{CASIA}\\\textbf{MFSD}    \end{tabular} & \multicolumn{1}{|c}{}                         \\ \hline
		\textbf{	Appearance information    }         & \multicolumn{2}{c|}{30.8\%} & \multicolumn{2}{c|}{49.7\%} &40.3\%                            \\ 
		\textbf{	Motion information   }     & \multicolumn{2}{c|}{15.6\%} & \multicolumn{2}{c|}{44.1\%}&  29.8\%                           \\ \hline
	\end{tabular}
\end{table}

\highlight{To explore the influence of optical flow information, we consider to use appearance information and motion information in the experiments, respectively.  In the appearance information situation, we only use the original frame from video, which is taken as the input of the proposed method. In the motion information situation, only the optical information is obtained from original videos. A performance comparison between these two cases is presented in Tab. \ref{crossdataset_ab}. Compared with appearance information, the motion information could better assist in distinguishing the spoofing faces from live faces. }

\subsection{Intra NUAA Database Experiment}

We evaluate the performance of the proposed method and state-of-the-art techniques on the NUAA dataset, in an intra-database sense. The competitors include DoG and high frequency based (DoG-F) \citep{li2004live}, multiple difference of Gaussian (DoG-M) \citep{ zhang2012face}, DoG-sparse logistic (DoG-SL) \citep{peixoto2011face}, diffused speed-local speed pattern (DS-LSP) \citep{ kim2015face}, multiple local binary pattern (M-LBP) \citep{ maatta2011face}, DoG-sparse low-rank bilinear logistic regression (DoG-LRBLR) \citep{Tan2010FaceLD}, DoG-sparse logistic (DoG-SL) \citep{peixoto2011face}, component-dependent descriptor (CDD) \citep{yang2013face}, ADKMM \citep{yu2017anisotropic} and the nonlinear diffusion based convolution neural network (ND-CNN) \citep{alotaibi2017deep}.

\begin{table}
\scriptsize
	\caption{Performance comparison using AUC on the NUAA dataset}
	\label{tab:nuaa}
	\centering
		\begin{tabular}{ll}
			\hline
			\textbf{Methods} & \textbf{Accuracy} \\
			\hline
			\textbf{Ours(semi-supervised)}    & 99.3\%    \\
			\textbf{ADKMM ('17)}            & 99.3\%            \\
			\textbf{ND-CNN ('17)}           & 99.3\%            \\
			\textbf{DS-LSP ('15)}           & 98.5\%            \\
			\textbf{CDD ('13)}             & 97.7\%            \\
			\textbf{DoG-SL ('11)}           & 94.5\%            \\
			\textbf{M-LBP ('11)}            & 92.7\%            \\
			\textbf{DoG-LRBLR ('10)}        & 87.5\%            \\
			\textbf{DoG-F ('04)}            & 84.5\%            \\
			\textbf{DoG-M ('12)}            & 81.8\%            \\
			\hline
		\end{tabular}
\end{table}

\begin{figure}[h]
	\centering
	\includegraphics[width=\linewidth,height=5cm]{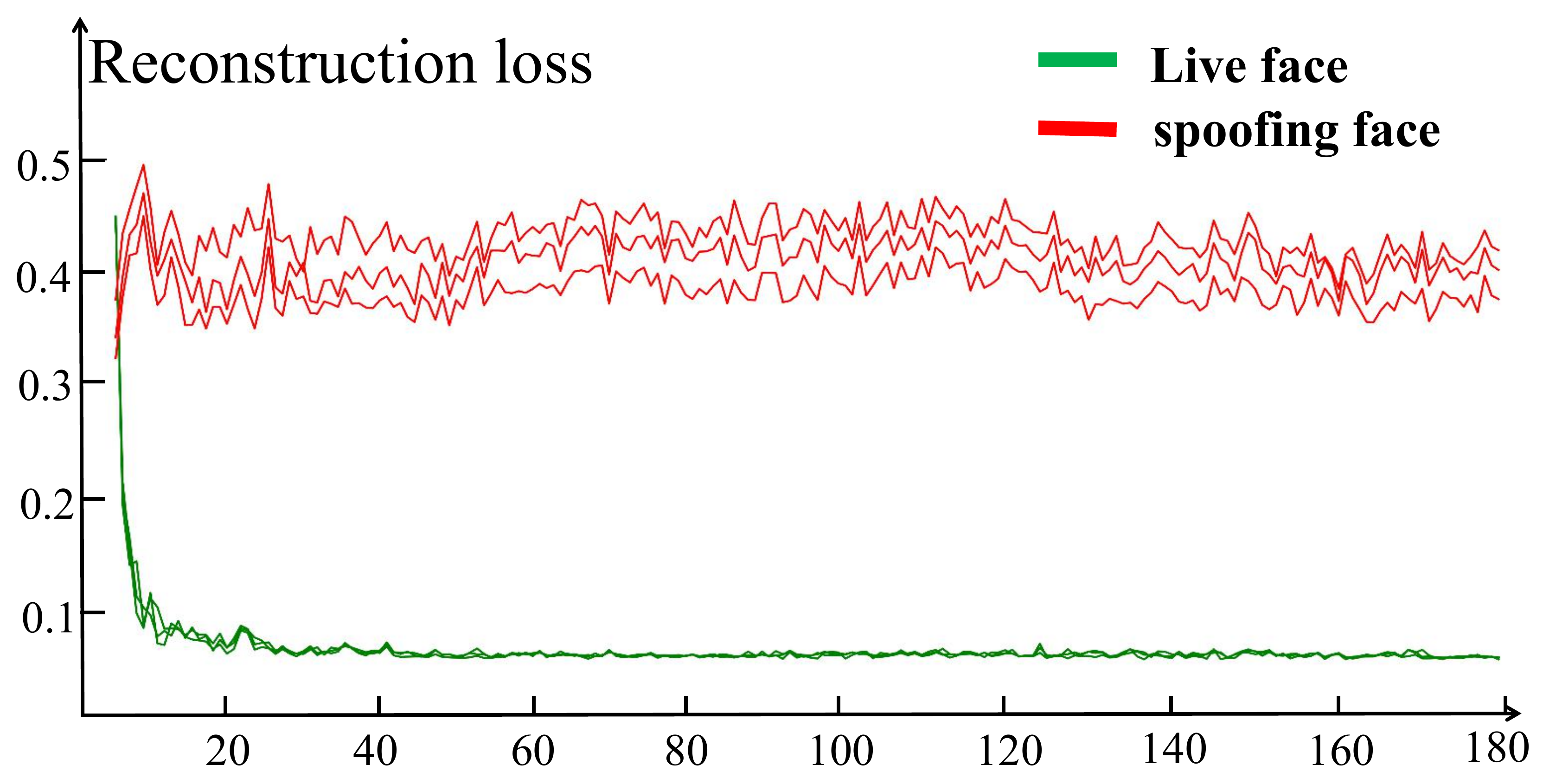}
	\caption{Reconstruction performance with different numbers of iterations (epoch) on the NUAA dataset. 
	}
	%  \Description{}
	\label{nuaa_loss}
\end{figure}

To evaluate the reconstruction performance for each epoch, we choose three face spoofing samples and three live samples from the train set randomly. Once the network are trained in each epoch, the trained model would output the reconstructed images of these live or spoofing samples. Fig. \ref{nuaa_loss} reveals that the gap between the reconstruction losses of spoofing faces and those of live faces are increased until they become stable after a few epochs. This indicates that the proposed approach can quickly distinguish spoof faces from live samples, without requiring spoof face data for training. Tab. \ref{tab:nuaa} shows the accuracies for all methods. %due to strong performance of the semi-supervised learning by adversarial training,
Our semi-supervised approach achieves the best performance, which is the same as the supervised ADKMM \citep{yu2017anisotropic} and supervised ND-CNN \citep{alotaibi2017deep}. %Most of live and spoof test samples can be accurately separated with an accuracy of 0.993. 

\subsection{Intra Replay-Attack Dataset Experiment}
\highlight{We also compared our method with state-of-the-art techniques on the Replay-Attack dataset, in the intra-database setting. Competitors includes
$\operatorname{LBP}_{3 \times 3}^{u 2}+x^{2}$ \citep{chingovska2012effectiveness}, $\mathrm{LBP}_{3 \times 3}^{u 2}+\mathrm{LDA}$ \citep{chingovska2012effectiveness}, $\mathrm{LBP}_{3 \times 3}^{\mathrm{u} 2}+\mathrm{SVM}$ \citep{chingovska2012effectiveness}, $\mathrm{LBP}+\mathrm{SVM}$ \citep{maatta2011face},
DS-LBP \citep{kim2015face}, ND-CNN \citep{alotaibi2017deep},
VGG \citep{li2016original}, Color-texture \citep{boulkenafet2015face},  Fisher-vector-encoding \citep{boulkenafet2016face}, Depth-based-CNNs (Patch-based CNN 
, Depth -based CNN and Patch and depth CNN) \citep{atoum2017face}, D-K \citep{yu2017anisotropic}, DTCNN \citep{tu2019deep}, Hand-crafted + CNN \citep{rehman2020enhancing} and Generalized deep feature \citep{li2018learning}. Previous spoofing face detectors have achieved outstanding performance in the intra-dataset setting by supervised learning with both positive and negative labels. To certain degree, these supervised methods have the risk of overfitting on the training data and obtain poor generalization in cross-dataset setting. In addition, in the real world, it is impossible for us to collect and cover all kinds of spoof faces. Some types of spoofing faces are even unknown. Based on these challenges, we formulate the spoofing faces detection task as an abnormal detection task by only training the normal samples (live faces), which obtain a comparable performance in intra dataset setting with strong generalization.} 

\begin{figure}[h]
	\centering
	\includegraphics[width=\linewidth]{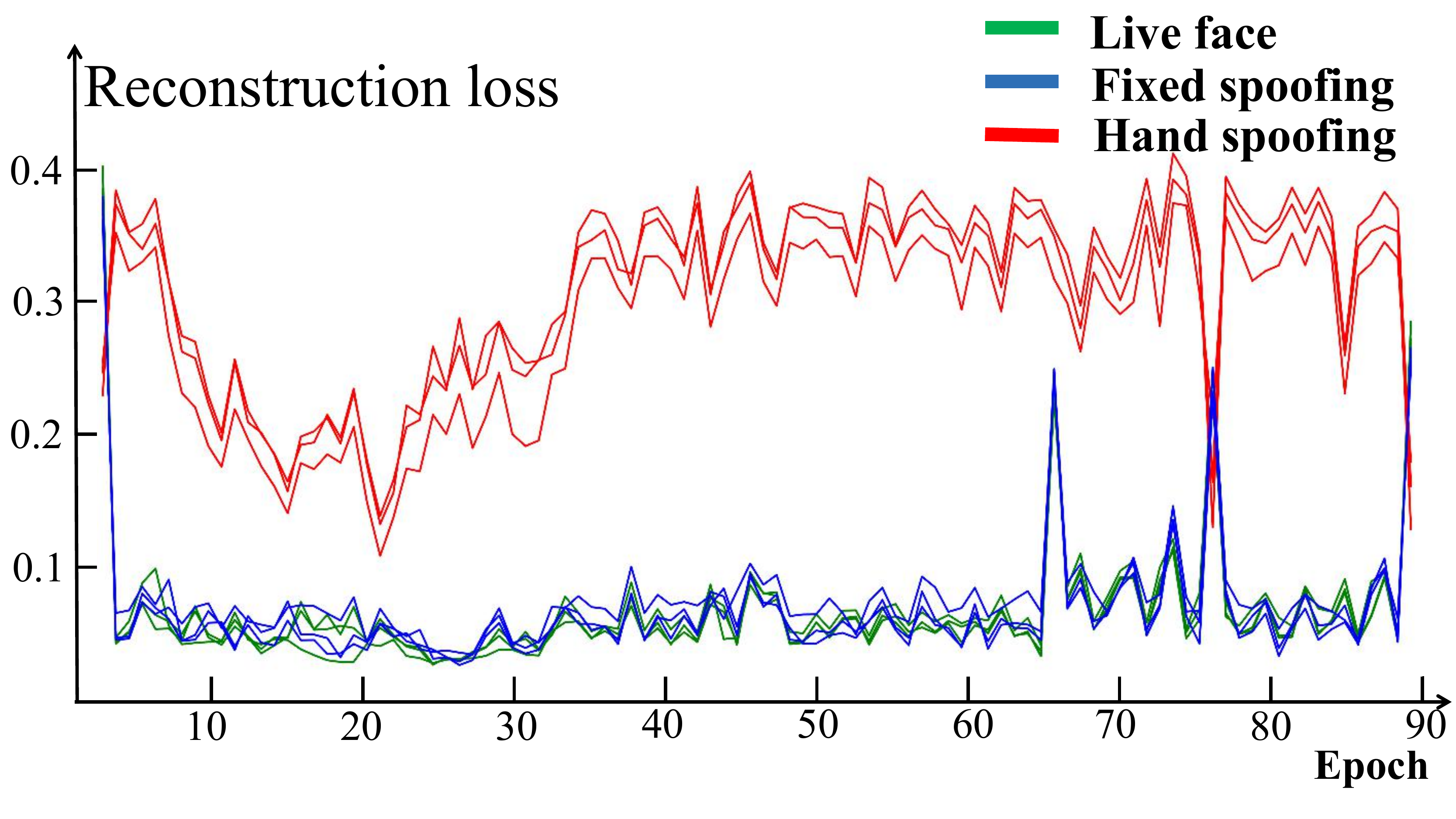}
	\caption{Three kinds of reconstruction performance with different iteration numbers (epoch) on the intra Replay-Attack database experiment.
	}
	%  \Description{}
	\label{replayattack_loss}
\end{figure}

The proposed method obtains the best performance with only $40$ epochs. Fig. \ref{replayattack_loss} shows the reconstruction loss for live and spoof faces. Besides the training data and testing data, this dataset also provides the development data to evaluate the performance. we calculate the half total error rate ($HTER$) \citep{ bengio2004statistical} to measure the performance. The $HTER$ is half of the sum of the false rejection rate ($FRR$) and false acceptance rate ($FAR$). The half total error rate ($HTER$) would be also used in the metric of cross-database experiments.

\begin{table}
\scriptsize
	\caption{\highlight{Performance comparison using HTER measure on the Replay-Attack dataset (intra-database setting).} }
	\label{tab:replay_attack}
	\centering
	\begin{tabular}{lll}
		\hline
		\textbf{Methods} & \textbf{test}\\
		\hline
		$\operatorname{\textbf{LBP}}_{3 \times 3}^{u 2}+x^{2}$ \textbf{('12)}     & 34.0\%            \\
		\textbf{$\mathrm{\textbf{LBP}}_{3 \times 3}^{u 2}+\mathrm{\textbf{LDA}}$ \textbf{('12)}}       & 17.2\%            \\
		\textbf{$\mathrm{\textbf{LBP}}_{3 \times 3}^{\mathrm{u} 2}+\mathrm{\textbf{SVM}}$ \textbf{('12)}}       & 15.16\%            \\
		\textbf{$\mathrm{\textbf{LBP}}+\mathrm{\textbf{SVM}}$ ('11)}             & 13.9\%            \\
		\textbf{DS-LBP ('15) }          & 12.5\%            \\
		\textbf{Color-texture ('15)} & 2.9\%            \\
		\textbf{VGG ('16)}           & 4.3\%            \\
        \textbf{D-K ('16)}           & 4.3\%            \\
		\textbf{Fisher-vector-encoding ('16)} & 2.0\%            \\
		\textbf{ND-CNN ('17) }          & 10.0\%            \\
		\highlight{\textbf{Patch-based CNN ('17)}}  &1.2\%  \\
		\highlight{\textbf{Depth -based CNN ('17)}} &0.7\%  \\
		\highlight{\textbf{Patch and depth CNN ('17)}} & 0.7\%  \\
		\highlight{\textbf{Generalized deep feature ('18)}} &1.2\%  \\
		\highlight{\textbf{DTCNN ('19)}} & 20.0\%    \\
		\highlight{\textbf{Hand-crafted + CNN ('20)}} & 2.3\%    \\
		\textbf{Ours}    & 12.3\%    \\
		\textbf{Ours with motion judgment }    & 3.5\%    \\
		\hline
	\end{tabular}
\end{table}

\begin{equation}
H T E R=\frac{F R R+F A R}{2}
\end{equation}

\begin{figure}[h]
	\centering
	\includegraphics[width=\linewidth]{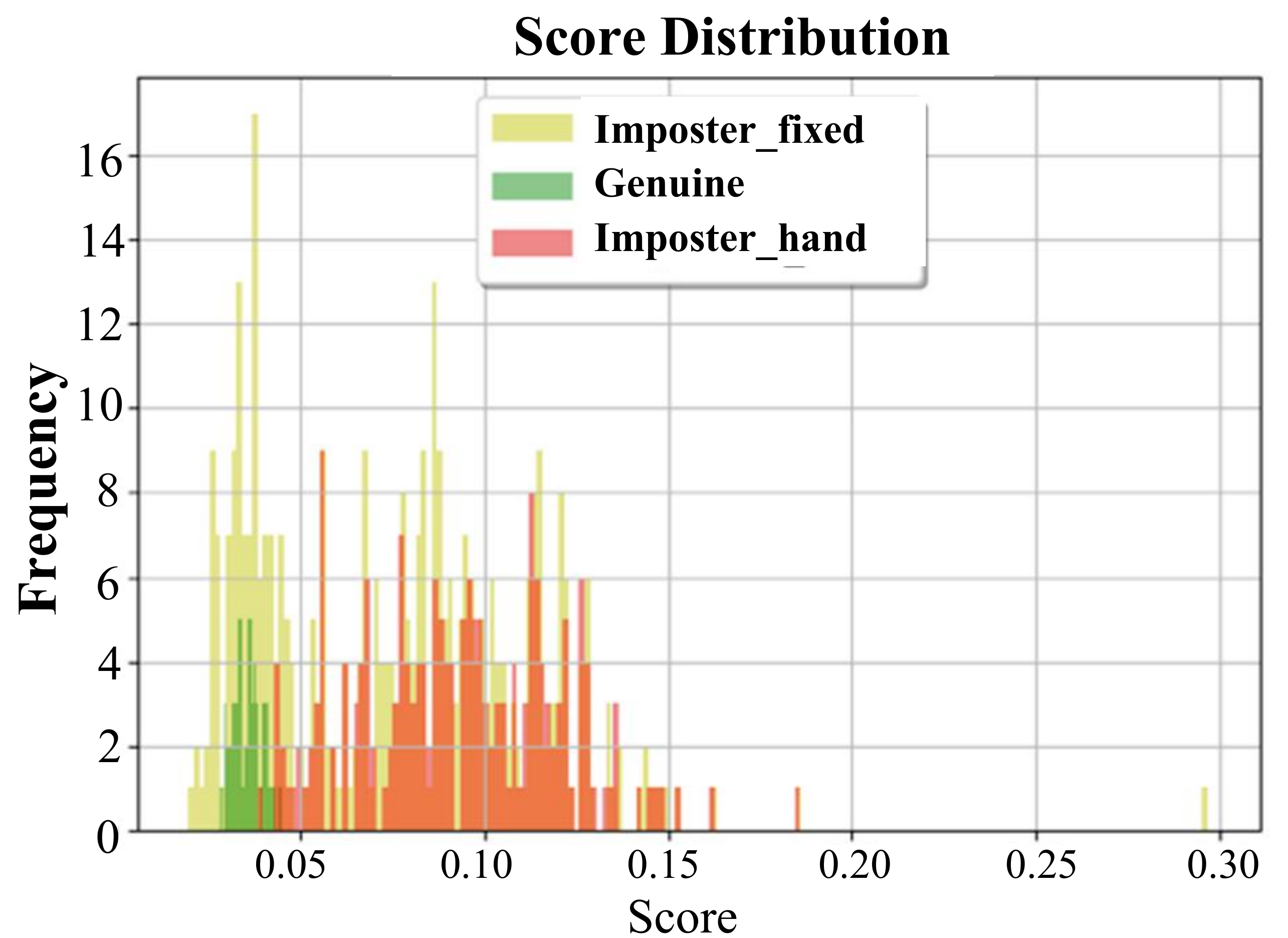}
	\caption{Three distributions of reconstruction error scores. It consists of live faces, spoofing faces by holding the client biometry (i.e., fixed spoofing) and spoofing faces from the device held by the attacker's hands (i.e., hand spoofing).  }
	%  \Description{}
	\label{replay_dis}
\end{figure}

The way to perform the attacks can be divided into two subsets: the first subset is composed of videos generated using a stand to hold the client biometry (``fixed''). For the second subset, the attackers hold the devices with their own hands. We choose three samples from each set (fixed spoofing, hand spoofing and live validation). Fig. \ref{replayattack_loss} illuminates that the reconstruction loss of live faces decreases sharply at the beginning of training. It tends to be stable with minor changes after that, until the $66$-th epoch where both the reconstruction losses of live and spoof faces start to oscillate.

It is noteworthy that the changes in live reconstruction error and fixed spoof faces reconstruction error are consistent with increasing iterations. From Fig. \ref{fixed_optical_flow}, there is no temporal information in the way of using a stand to hold the client biometry (``fixed'') except some noise in the optical flow maps. Thus, the performance of reconstruction loss for fixed spoofing faces would be great and even better than live faces. Fig. \ref{replay_dis} also further verifies what we have found in Fig. \ref{fixed_optical_flow}.

A performance comparison with previous methods is shown in Tab. \ref{tab:replay_attack}. On the test set of the Replay-Attack dataset, the $HTER$ of our method is $0.123$, and we achieve a comparable HTER to other methods (worse than the best). This result attributes to two factors. Firstly, the binary classification methods, using both positive and negative data with labels, often achieve excellent performance in the intra-database setting (i.e., train and test within the same dataset). Some of the compared methods even use depth information or other extra cues for spoof faces detection. Secondly, as we explained before, most fixed spoof faces are mistakenly identified as live faces based on low reconstruction loss due to no motion cues (e.g., Fig. \ref{fixed_optical_flow}).

To tackle this issue, we design a new module which is responsible for detecting the presence of motion information. This is achieved through calculating the average pixel difference between pairs of optical flow maps of random frame samples. If there is no noticeable difference between each pair of optical flow maps, it means no motion information. With the support of this motion detection module, we could obtain more information for spoof faces before the real face spoofing detection. The $HTER$ of our method can be declined from $0.123$ to $0.035$, with the aid of such motion judgment.

\begin{figure*}[t!]
	\centering
	\includegraphics[width=\linewidth,height=6cm]{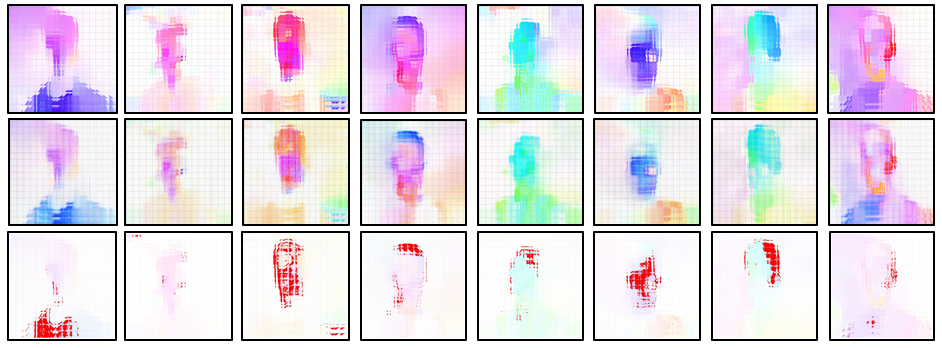}
	\caption{First row: original optical flow images of live faces. \textcolor{black}{Second row: generated optical flow images of live faces. Third row: corresponding maps which display the differences between the original images (the first row) and generated images (the second row) from model by red points.} }
	%  \Description{}
	\label{live_reconstruction}
\end{figure*}

\begin{figure*}[t!]
	\centering
	\includegraphics[width=\linewidth,height=6cm]{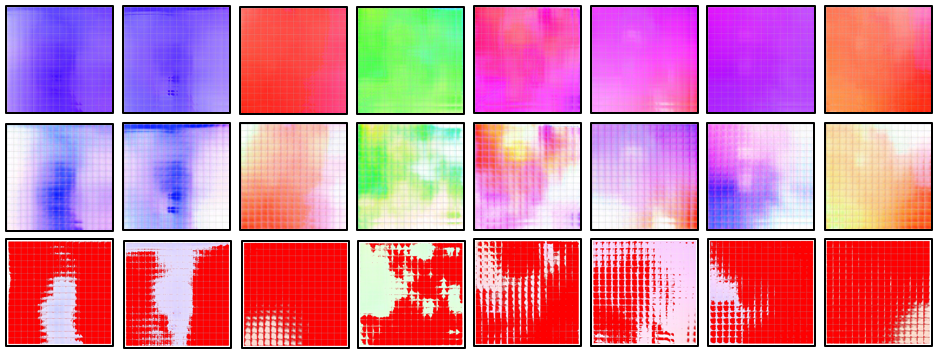}
	\caption{First row: original optical flow images of spoofing faces. Second row: generated optical flow images of spoofing faces. Third row: corresponding maps which display the differences between original image (the first row) and generated images (the second row) by red point.}
	%  \Description{}
	\label{spoofingbyhandreconstruction}
\end{figure*}

\begin{figure}[h]
	\centering
	\includegraphics[width=\linewidth]{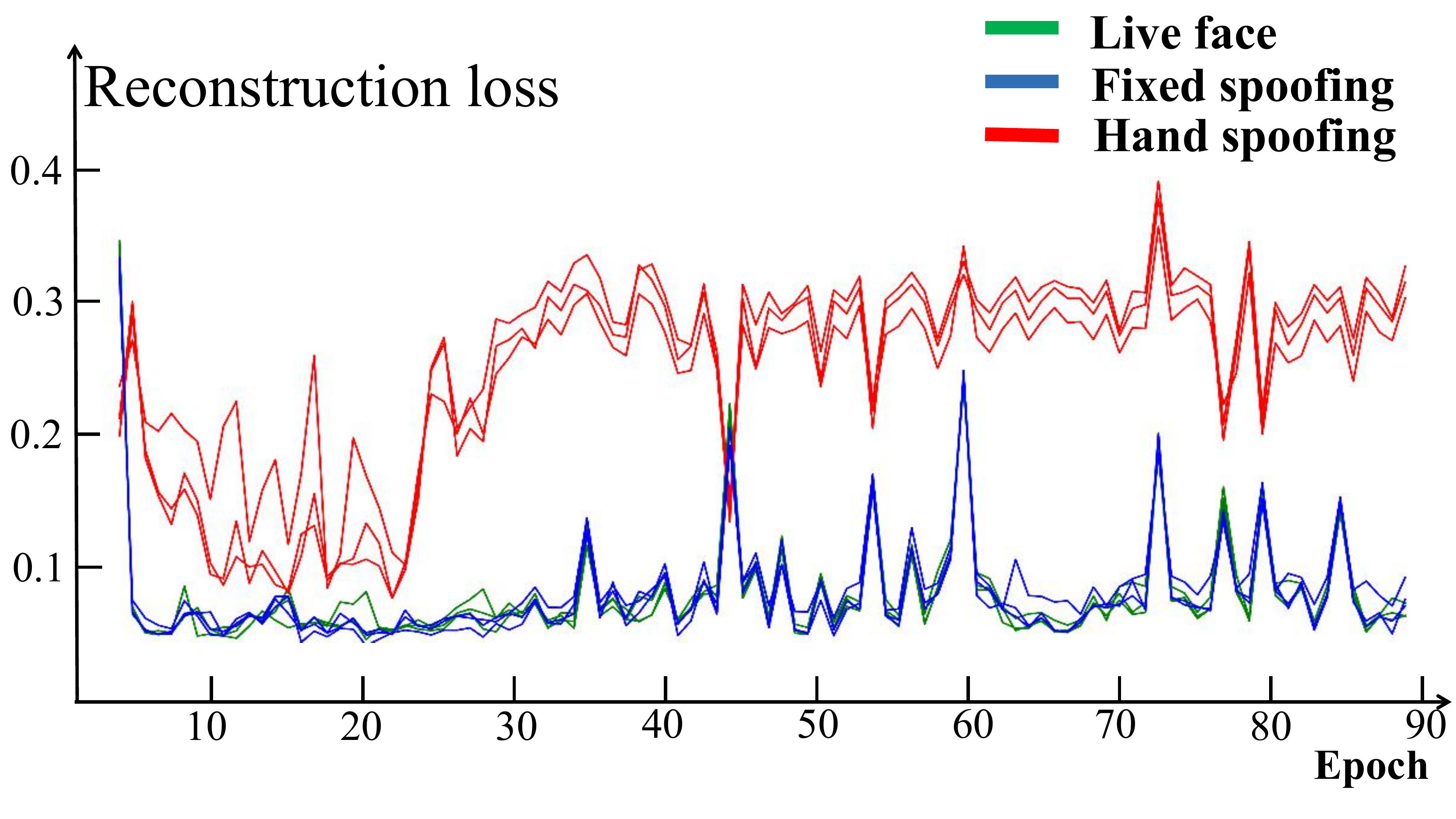}
	\caption{Three kinds of reconstruction performance with different iteration numbers (epoch) on cross-database experiment with training on the training set of the CASIA-MFSD database and testing on the testing set of the Replay-Attack database. }
	%  \Description{}
	\label{cross_loss}
\end{figure}

\begin{figure}[h]
	\centering
	\includegraphics[width=\linewidth]{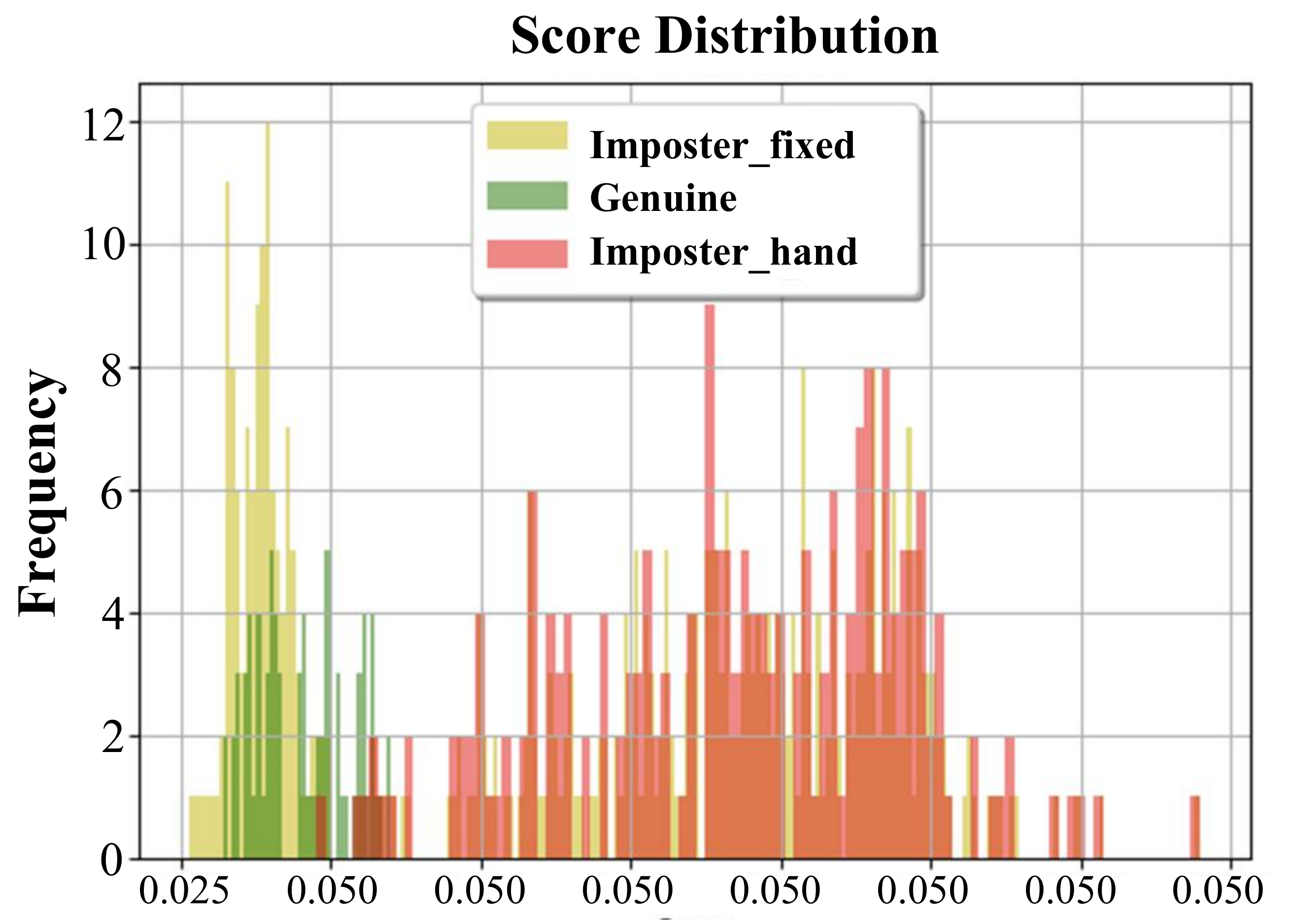}
	\caption{Three distributions of reconstruction error scores in experiment with training on the training set of the CASIA-MFSD database and testing on the testing set of the Replay-Attack database. }
	%\Description{}
	\label{cross_dis}
\end{figure}

\subsection{Cross-database Experiments}

The cross-database performance is evaluated by training the proposed method on the CASIA-MFSD dataset and testing it on the Replay-Attack dataset, and vice versa.

\begin{table}[h]
	\scriptsize
	\caption{\highlight{Classification performance in terms of HTER (\%). The models are trained using the CASIA-MFSD dataset and tested on the Replay-Attack dataset, and vice versa. 1: supervised method. 2: semi-supervised method.} }
	\label{crossdataset}
	\begin{tabular}{c|c|c|c|c|c}
		\hline
		\multicolumn{1}{c|}{\multirow{2}{*}{\textbf{Methods}}} & \multicolumn{1}{l|}{\textbf{Train}}                                                & \textbf{Test}& \multicolumn{1}{l|}{\textbf{Train}} & \textbf{Test}                                                    & \multicolumn{1}{|c}{\multirow{2}{*}{\textbf{Average}}} \\ \cline{2-5}
		& \multicolumn{1}{l|}{\begin{tabular}[c]{@{}l@{}}\textbf{CASIA}\\\textbf{MFSD}\end{tabular}} & \begin{tabular}[c]{@{}l@{}}\textbf{Replay}\\\textbf{Attack}\end{tabular} &
		\multicolumn{1}{l|}{\begin{tabular}[c]{@{}l@{}}\textbf{Replay}\\\textbf{Attack} \end{tabular}} & \begin{tabular}[c]{@{}l@{}} \textbf{CASIA}\\\textbf{MFSD}    \end{tabular} & \multicolumn{1}{|c}{}                         \\ \hline
		\textbf{	1-LBP ('13)  }          & \multicolumn{2}{c|}{47.0\%}& \multicolumn{2}{c|}{39.6\%}& 43.3\%                                       \\ 
		\textbf{	1-LBP-TOP ('13)     }             & \multicolumn{2}{c|}{49.7\%}& \multicolumn{2}{c|}{60.6\%}& 55.2\%   \\ 
		\textbf{	1-Motion  ('13)      }      & \multicolumn{2}{c|}{50.2\%}& \multicolumn{2}{c|}{47.9\%} & 49.1\%                                       \\
		\textbf{	1-CNN  ('14)    }  & \multicolumn{2}{c|}{48.5\%}& \multicolumn{2}{c|}{45.5\%}& 47.0\%                                       \\ 
		\textbf{	1-Color LBP  ('15)    }     & \multicolumn{2}{c|}{37.9\%}& \multicolumn{2}{c|}{35.4\%}& 36.7\%                                        \\
		\textbf{	1-Color Tex  ('16)  }   & \multicolumn{2}{c|}{30.3\%} & \multicolumn{2}{c|}{37.7\%}& 34.0\%                                          \\ 
		\textbf{	1-Auxiliary('18)   }   & \multicolumn{2}{c|}{27.6\%}& \multicolumn{2}{c|}{\textbf{28.4\%}}& \textbf{28.0}\%                                          \\ 
		\textbf{	1-De-Spoof('18)   }     & \multicolumn{2}{c|}{28.5\%}& \multicolumn{2}{c|}{41.1\%}& 34.8\%                                          \\ 
		
				\highlight{\textbf{	1-DA ('18)    } }        & \multicolumn{2}{c|}{27.4\%} & \multicolumn{2}{c|}{36.0\%} &31.7\%                            \\ 
		
		\textbf{	1-Dynamic texture ('18)    }         & \multicolumn{2}{c|}{22.2\%} & \multicolumn{2}{c|}{35.0\%} &28.6\%                            \\

		\highlight{\textbf{	1-OF Domain ('18)    }}         & \multicolumn{2}{c|}{30.1\%} & \multicolumn{2}{c|}{36.8\%} &33.5\%                            \\ 
		
		\highlight{\textbf{	1-GFA-CNN ('19)    }}         & \multicolumn{2}{c|}{21.4\%} & \multicolumn{2}{c|}{34.3\%} &28.0\%                            \\

		\highlight{\textbf{	1-ADA ('19)    }}         & \multicolumn{2}{c|}{17.5\%} & \multicolumn{2}{c|}{41.6\%} &29.6\%                            \\

		\textbf{	2-Proposed method   }     & \multicolumn{2}{c|}{\textbf{15.6\%}} & \multicolumn{2}{c|}{44.1\%}&  29.8\%                            \\ \hline
	\end{tabular}
\end{table}

We first consider training on the training set of the CASIA-MFSD database and testing on the testing set of the Replay-Attack database. The quantitative results shown in Tab. \ref{crossdataset} confirm that the proposed method achieves the best performance ($HTER=0.156$) on the Replay-Attack test set which includes different types of spoofing attacks. \highlight{The competitors consist of LBP \citep{de2013can},
LBP-TOP \citep{de2013can},
Motion \citep{de2013can},
CNN  \citep{yang2014learn},
Color LBP \citep{boulkenafet2015face},
Color Tex \citep{boulkenafet2016face},
Auxiliary \citep{liu2018learning},
De-Spoof \citep{jourabloo2018face},DA \citep{li2018unsupervised},Dynamic texture \citep{shao2018joint}, OF Domain \citep{sun2018investigation}, ADA \citep{wang2019improving} and GFA-CNN \citep{tu2019learning}.} In Fig. \ref{cross_dis}, it is obvious that the trained model has strong generalization ability to make live faces and fake faces obtained by using the attackers' bare hands separable. However, some fake faces obtained by fixed support has the same distribution as live faces.

We then conduct the opposite experiment: training on the training set of the Replay-Attack dataset and testing on the testing set of the CASIA-MFSD dataset. Our method achieves a competitive performance ($HTER=0.441$) for the cross testing on the testing set of the CASIA-MFSD dataset. From Tab. \ref{crossdataset}, we can see that the $HTER$ of our method is better than most binary supervision methods \citep{yang2014learn,jourabloo2018face,wu2016motion,boulkenafet2016face,boulkenafet2015face}. This demonstrates that the proposed approach can better identify the differences between live and fake faces.

As with all previous works (Wu et al. 2016; Jourabloo, Liu, and Liu 2018; Boulkenafet, Komulainen, and Hadid 2016), we observe that the models trained on CASIA-MFSD enables to generalize better than the model trained on the Replay Attack Database. We speculate as follows (1) It is probably because the resolution of the CASIA-MFSD data is significantly higher than that in the Replay-Attack dataset. The model trained with high resolution could generalize better than the model trained with low resolution. (2) Compared with Replay-Attack, the CASIA-MFSD contains more variations in collected database, For example, imaging quality, the distance between camera and face, background and attack types. Hence, the model optimized for Replay-Attack databases faces more challenge in the new acquisition con- ditions. This is one limitation of the our method and previous works, and worthy further research. As with previous works \citep{wu2016motion,jourabloo2018face,boulkenafet2016face}, the two above cross-database experiments do not have the same $HTER$.  We speculate that it is probably because the resolution of the CASIA-MFSD data is significantly higher than that in the Replay-Attack dataset. In Fig. \ref{live_reconstruction} and \ref{spoofingbyhandreconstruction}, the first row and second row show the optical flow maps and the reconstructed optical flow maps, respectively. The third row visualizes the differences between the corresponding optical maps and reconstructed maps (red color). These visualization results clearly demonstrate that the reconstruction errors from live faces are lower than that of spoofing faces.

%% file: paper/conclusion.tex
\section{Conclusion}
\label{sec:conclusion}
We have presented an adversarial framework for the detection of spoofing faces. Given an input face image, the trained model can automatically determine if it is a live or spoof face. Current face anti-spoofing techniques have to utilize both spoof data and live data for training, which can hardly cover every type of spoof faces. By contrast, our approach does not need spoof data for training, and is thus semi-supervised and robust to different types of spoof faces. Both the intra-/cross-database experiments show that our method achieves better or comparable results to state-of-the-art techniques. We believe our research will arouse some new insights in this field.